\documentclass{article}

\usepackage{arxiv}

\usepackage[utf8]{inputenc} 
\usepackage[T1]{fontenc}    
\usepackage{hyperref}       
\usepackage{url}            
\usepackage{booktabs}       
\usepackage{amsfonts}       
\usepackage{nicefrac}       
\usepackage{microtype}      
\usepackage{lipsum}

\usepackage{calrsfs,mathtools,amsmath,amsfonts,amssymb,amsthm,bbm,mathbbol,etoolbox}
\usepackage{cancel}
\DeclareMathAlphabet{\pazocal}{OMS}{zplm}{m}{n}
\newcommand{\CJ}{\pazocal{J}}

\DeclarePairedDelimiterX{\abs}[1]{\lvert}{\rvert}{\ifblank{#1}{{}\cdot{}}{#1}}

\title{A semi-supervised sparse K-Means algorithm}

\author{
  Avgoustinos ~Vouros \\
  Department of Computer Science\\
  University of Sheffield\\
  Sheffield S10 2TN, United Kingdom \\
   \And
 Eleni ~Vasilaki \\
  Department of Computer Science\\
  University of Sheffield\\
  Sheffield S10 2TN, United Kingdom \\
  \texttt{e.vasilaki@sheffield.ac.uk} \\
}

\begin{document}
\maketitle

\begin{abstract}
We consider the problem of data clustering with unidentified feature quality and when a small amount of labelled data is provided. An unsupervised sparse clustering method can be employed in order to detect the subgroup of features necessary for clustering and a semi-supervised method can use the labelled data to create constraints and enhance the clustering solution. In this paper we propose a K-Means variant that employs these techniques. We show that the algorithm maintains the high performance of other semi-supervised algorithms and in addition preserves the ability to identify informative from uninformative features. We examine the performance of the algorithm on synthetic and real world data sets. We use scenarios of different number and types of constraints as well as different clustering initialisation methods.
\end{abstract}


\section{Introduction}

In many learning tasks we have a plethora of unlabelled data in a high-dimensional space, where coordinates correspond to a series of features. We are also provided with a limited number of labelled data since the latter is expensive to generate. In many cases we do not have knowledge of the actual contribution of each feature on the learning task and we often consider dimensionality reduction methods in order to keep only the most relevant features to the given task. However, many dimensionality reduction methods, such as Principal Component Analysis \cite{wold1987principal}, result in a transformation of the original features which limits their interpretability, especially when features were designed to have a physical meaning. In unsupervised scenarios a number of authors \cite{wang2008variable,xie2008penalized,raftery2006variable,maugis2009variable} have proposed clustering algorithms that have the ability to keep the initial features intact and assign a certain weight to them based on their contribution to the clustering. These algorithms result in feature selection and sparse clustering. 

\cite{witten2010framework} present a general sparse clustering framework which incorporates $L_1$ (Lasso regression) and $L_2$ (Ridge regression) penalties to eliminate the uninformative features and provide weights the rest according to their contribution to the clustering \cite{witten2009penalized}. Such method requires the tuning of the \textit{sparsity} hyper-parameter which regulates the $L_1$ effect. This framework has been applied on K-Means and Hierarchical clustering but can also be applied to semi-supervised scenarios where pairwise constraints are given as an additional input to the algorithm. Such constraints are generated from partly labelled data and indicate which data points should (MUST-LINK) or should not (CANNOT-LINK) belong to the same cluster. Previous work on semi-supervised learning \cite{wagstaff2001constrained,basu2002semi,klein2002instance,xing2003distance,bar2003learning} shows that incorporated constraints can result in a performance improvement for the learning algorithm. This is achieved by guiding the clustering solution either by modifying the objective function of the algorithm to include satisfaction of the constraints \cite{demiriz1999semi} or by initialising the centroids at more appropriate locations of the feature space based on the constraints \cite{basu2002semi}. Another technique is to train a metric that satisfy the constraints as in the study of \cite{xing2003distance}.

Based on the work of \cite{witten2010framework} on sparse K-Means clustering we propose a modification to the objective function of their algorithm to incorporate constraints. We show that in this case we get the best of both worlds since constraints result in better clustering performance without affecting the sparsity capabilities of the algorithm. We name this algorithm Pairwise Constrained Sparse K-Means (PCSKM) and we test its performance under different conditions such as different number and kind of constraints (CANNOT-LINK, MUST-LINK or both). In our previous study \cite{vouros2019empirical} we have shown that the deterministic initialisation method of Density K-Means++ (DKM++) \cite{nidheesh2017enhanced} surpasses the average performance of stochastic methods thus we select this method for the initialisation of the algorithms along with the Seeding method proposed in the study of \cite{bilenko2004integrating}. We have also included the initialisation methods of ROBIN \cite{al2009robust} and Maximin \cite{katsavounidis1994new} to strengthen our conclusions (results in appendix). In our benchmark we include synthetic data sets from the study of \cite{brodinova2017robust} with known feature quality, real world data sets from the UCI repository \cite{asuncion2007uci} and two real world data sets from our previous behavioural neuroscience study \cite{vouros2018generalised} which contain 10 known uninformative features.

The data sets used in this study, including their constraints, and the MATLAB code to run the simulations are available on the GitHub repository \href{https://github.com/avouros/Code-PCSKM}{\text{avouros/Code-PCSKM}}.

\section{Methods}

We will use the following notations and assumptions:
\begin{itemize}
	\item A data set is presented in the form of $n$ by $p$ matrix, where $n$ is the number of observations and $p$ is the number of features (dimensions), and $x_{i:} = [x_{i1}, x_{i2}, \dots, x_{ip}]$ specifies the i-th element of the data set.
	\item $K$ represents the number of target clusters $\pazocal{C} = \{c_1, c_2, \dots, c_K \}$ with $n_1, n_2, \dots, n_K$ number of elements in each cluster respectively (it is assumed that the clusters may not be of equal sizes). The $m_{k:} = [m_{k1}, m_{k2}, \dots, m_{kp}]$ specifies the k-th cluster center (centroid). The centroid is the mean of the data points in this cluster.
	\item ${\mu}_{1:} = {\mu}_{11}, {\mu}_{12}, \cdots, {\mu}_{1p}$ is the global centroid of the data set (mean of all data).
	\item $\CJ$ will be used for functions intended for minimization and $J$ for functions intended for maximization.
	\item The distance metric in all the algorithms is the squared Euclidean distance. 
\end{itemize}

\subsection{The K-Means (LKM) algorithm}

K-Means aims to minimize the sum of squared distances between the data points and their respected centroids as indicated in the objective function: 
\begin{align}\label{AlgoKMeans}
\CJ_{kmeans} &= \sum_{k=1}^{K}\sum_{\binom{i=1}{x_{i:} \in c_k}}^{n_k} \sum_{j=1}^{p} (x_{ij}-m_{kj})^2	
\end{align}	
which is equivalent to maximizing the between cluster sum of squares (BCSS) given by equation:
\begin{align}\label{BCSS}
J_{BCSS} &= \sum_{i=1}^{n}\sum_{j=1}^{p}(x_{ij}-{\mu}_{1j})^2 - \sum_{k=1}^{K}\sum_{\binom{i=1}{x_{i:} \in c_k}}^{n_k}\sum_{j=1}^{p}(x_{ij}-m_{kj})^2	
\end{align}	

The most common algorithm to minimize equation (\ref{AlgoKMeans}) is the Lloyd's K-Means algorithm which
is described below \cite{jain2010data}:
\begin{enumerate}
	\item Initialise $K$ initial centroids $M = \{m_{1j}, \dots, m_{Kj}\}$ using an initialisation method. 
	\item Assign each data point to cluster $k^*$ so that,
	\begin{align}\label{lkmeansobj}
	k^* &= \underset{k}{argmin}\bigg\{ \sum_{\binom{i=1}{x_{i:} \in c_k}}^{n_k}  \sum_{j=1}^{p} (x_{ij}-m_{kj})^2    \bigg\}
	\end{align}	
	\item Recompute the cluster centroids by taking the mean of the data points assigned to them, i.e for the k-th cluster, if it contains $n_k$ elements its centroid is computed as $m_{kj} = \frac{\sum_{\binom{i=1}{x_{i:} \in c_k}}^{n_k} x_{ij}}{n_k}$
	\item Repeat steps 2 and 3 until converge, i.e. there are no more data points reassignments.						
\end{enumerate}	
The algorithm returns the final clusters (centroids and element assignments).

\subsection{The Sparse K-Means (SKM) algorithm}

In the work of \cite{witten2010framework} the authors propose the maximisation of a weighted version of the BCSS subject to certain constraints. The proposed objective function is given by:
\begin{align}\label{skmeans}
J_{skmeans} =& \sum_{i=1}^{n}\sum_{j=1}^{p} w_{j}(x_{ij}-{\mu}_{1j})^2 - \sum_{k=1}^{K}\sum_{\binom{i=1}{x_{i:} \in c_k}}^{n_k}\sum_{j=1}^{p} w_{j}(x_{ij}-m_{kj})^2 \\ \nonumber
\text{subject to}& \hphantom{xxxxx} \sum_{j=1}^{p} w_{j}^{2} \leq 1 \text{,}\hphantom{x} \sum_{j=1}^{p} \abs{w_{j}} \leq s \text{,}\hphantom{x} w_{j} \geq 0\hphantom{x} \forall j 
\end{align}	
where $\sum_{j=1}^{p} w_{j}^{2} \leq 1$ is the $L_2$ penalty or Ridge regression and $\sum_{j=1}^{p} \abs{w_{j}} \leq s$ is the $L_1$ penalty or Lasso regression. The minimization of the $L_1$ penalty will result in a constant shrinkage of the weights meaning that some weights will reach $0$ (feature selection), while the minimization of the $L_2$ penalty will result in proportional shrinkage of the weights (feature weighting). The parameter $s$ is known as the \textit{sparsity parameter} and regulates the amount of spareness, i.e. how many weights will become 0. A large weight value means that the feature contributed greatly to the clustering while a 0 weight means no contribution. An iterative algorithm for maximizing (\ref{skmeans}) is given by \cite{witten2010framework}:

\begin{enumerate}
	\item Initialise $K$ initial centroids $M = \{m_{1j}, \dots, m_{Kj}\}$ using an initialisation method. The authors propose $w_{1} = \dots = w_{p} = \frac{1}{\sqrt{p}}$.
	\item Holding the weights fixed, maximize (\ref{skmeans}) with respect to $M$. This can be implemented by performing K-Means on the scaled data i.e. multiply each feature $j$ with $\sqrt{w_j}$.
	\item Holding $M$ fixed optimize (\ref{skmeans}) with respect to the weights, which, according to \cite{witten2009penalized} leads to:
	
	\begin{align}\label{skmeansopt}
	w_{j} = \frac{sign(\gamma_{j})(\vphantom{x} \abs{\gamma_{j}}-\Delta)_{+}}{
		\sqrt{\sum_{j'=1}^{p}\big{(}sign(\gamma_{j'})(\vphantom{x} \abs{\gamma_{j'}}-\Delta)\big{)}^2}
	}
	\end{align}	
	\begin{align}\label{skmeansopt2}
	\gamma_j = \sum_{i=1}^{n} (x_{ij}-{\mu}_{1j})^2 - \sum_{k=1}^{K}\sum_{\binom{i=1}{x_{i:} \in c_k}}^{n_k}(x_{ij}-m_{kj})^2
	\end{align}	
	where $x_+$ denotes the positive part of $x$. We assume that $\gamma_j$ has a unique maximum and that $1 \leq s \leq \sqrt{p}$. Choose $\Delta = 0$ if that leads to $\sum_{j=1}^{p} \abs{w_{j}} \leq s$, otherwise find $\Delta > 0$ that results in $\sum_{j=1}^{p} \abs{w_{j}} = s$. To find $\Delta$ we can use the bisection method, see also \cite{witten2010framework}.
	
	\item Iterate through steps 2 and 3 until the convergence criterion in equation (\ref{skmeanconv})
	\begin{align}\label{skmeanconv}
	\frac{ \sum_{j=1}^{p}\abs{w_{j}^{r}-w_{j}^{r-1}} }{ w_{j}^{r-1} } < \epsilon
	\end{align}	
	where $w_{j}^{r}$ refers to the weights of the current iteration, $w_{j}^{r-1}$ to the weights of the previous iteration and $\epsilon = 10^{-4}$.	
\end{enumerate}	
The algorithm returns the final clusters (centroids and element assignments) and the weight of each feature.		

\subsection{The Pairwise Constrained K-Means (PCKM) and the Metric Pairwise Constrained K-Means (MPCKM) algorithms}	

In the study of \cite{bilenko2004integrating} the authors proposed the semi-supervised algorithm Metric Pairwise Constrained K-Means (MPCKM) which learns a distance metric based on constraints imposed by labelled data points. The constraints are imposed between pairs of points and can be either MUST-LINK, i.e. the two points must be in the same cluster or CANNOT-LINK, i.e. the two points must not be in the same cluster \cite{wagstaff2001constrained}. MPCKM integrates a variant of the Pairwise Constrained K-Means (PCKM) algorithm \cite{basu2004active} with metric learning \cite{bar2003learning,xing2003distance}. The PCKM algorithm \cite{basu2004active} incorporates constraints to guide the clustering solution; the constraints are considered as ``soft'', meaning that violations are permitted as opposed to its predecessor, the COP-KMeans algorithm \cite{wagstaff2001constrained},  which stops if constraints violation is unavoidable. Metric learning is the adaptation of a distance metric to satisfy the similarity imposed by the pairwise constraints. 

A key ingredient of both of PCKM and MPCKM is an initialisation process, known as Seeding, see \cite{basu2002semi}, which it is shown to improve the results of the algorithms on standard benchmarks. In this work, we consider separately the initialisation method from the clustering algorithm. PCKM considers constraints for clustering assignments without doing metric learning; MPCKM considers constraints for clustering assignments and performs metric learning using a single metric parametrized by a diagonal matrix for all the clusters, similar to MK-Means in \cite{bilenko2004integrating}. When Seeding is used these algorithms are the same as the ones described in \cite{bilenko2004integrating}. We should also note that, by default, MPCKM uses a single metric for each cluster but it also allows the use of full matrices, one for each cluster. These scenarios were not considered in this study since it is computationally expensive to calculate one metric per cluster \cite{bilenko2004integrating} and likely prone to overfitting. The use of a full matrix results in feature generation, i.e. the new features are linear combinations of the original ones \cite{xing2003distance}, which is outside the scope of this study.

The PCKM objective function is given by:
\begin{align}\label{pckmeans}
\CJ_{pckm} &= \sum_{k=1}^{K}  \sum_{\binom{i=1}{x_{i:} \in c_k}}^{n_k} \bigg( \sum_{j=1}^{p} (x_{ij}-m_{kj})^2 + 
\nonumber\\&
\sum_{(x_{i:})ML(x_{i':})} \sum_{j=1}^{p} b_{x_i,x_{i'}} f(x_{i:},x_{i':}) \hphantom{x}\mathbbm{1} \big[(x_{i:})\cancel{ML}(x_{i':})\big] + 
\nonumber\\&
\sum_{(x_{i:})CL(x_{i':})} \sum_{j=1}^{p} \bar{b}_{x_i,x_{i'}} \bar{f}(x_{i:},x_{i':}) \hphantom{x}\mathbbm{1} \big[(x_{i:})\cancel{CL}(x_{i':})\big] \bigg)
\end{align}	
\noindent where the second and third terms of the equation are two functions that indicate the severity of violating the imposed MUST-LINK and CANNOT-LINK constraints of the i-th element belonging to the k-th cluster; $\mathbbm{1}$ is a boolean function that specifies if in case of a MUST-LINK (${(x_{i:})ML(x_{i':})}$) or CANNOT-LINK (${(x_{i:})CL(x_{i':})}$) constraint, this constraints has been violated; $\big[(x_{i:})\cancel{ML}(x_{i':})\big]$ specifies violation of a MUST-LINK constraint and $\big[(x_{i:})\cancel{CL}(x_{i':})\big]$ violation of a CANNOT-LINK constraint. The terms $b_{x_i,x_{i'}}$ and $\bar{b}_{x_i,x_{i'}}$ are providing a way of specifying individual costs for each constraint violation. In the original algorithm by \cite{basu2004active} the functions $f(x_{i:},x_{i':})$ and $\bar{f}(x_{i:},x_{i':})$ were equal to $1$. Specifying appropriate values for constraint costs can be challenging and requires extensive knowledge about the data set under analysis or the constraints quality. In this study, similarly to \cite{bilenko2004integrating}, we assume: $b_{x_i,x_{i'}}=1$, $\bar{b}_{x_i,x_{i'}}=1$, $f(x_{i:},x_{i':}) = (x_{i:}-x_{i':})^2$ and $\bar{f}(x_{i:},x_{i':}) = \big((x_{I:}-x_{I':})^2 - (x_{i:}-x_{i':})^2\big)$, where $x_{I:}$, $x_{I':}$ are the maximally separated points in the data set. In this way the severity is proportional to the distance of $x_{I:}$ and $x_{I':}$. For $f(x_{i:},x_{i':})$ the severity of the penalty for violating a MUST-LINK constraint between $x_{i:}$ and another distant point $x_{i'j}$ is higher than a pair of nearby points. Analogously, for $\bar{f}(x_{i:},x_{i':})$, the severity of the penalty for violating a CANNOT-LINK constraint between $x_{i:}$ and another nearby point is higher than a pair of distant points. For minimising (\ref{pckmeans}) a variant of the K-Means algorithm is used which considers constraints in the data point assignment to the nearest cluster $k_{pckm}^*$:

\begin{align}\label{lpckmeansobj}
k_{pckm}^* &= \underset{k}{argmin}\bigg\{ \sum_{\binom{i=1}{x_{i:} \in c_k}}^{n_k} \bigg( \sum_{j=1}^{p} (x_{ij}-m_{kj})^2 + 
\nonumber\\&
\sum_{(x_{i:})ML(x_{i':})} \sum_{j=1}^{p} b_{x_i,x_{i'}} f(x_{i:},x_{i':}) \hphantom{x}\mathbbm{1} \big[(x_{i:})\cancel{ML}(x_{i':})\big] + 
\nonumber\\&
\sum_{(x_{i:})CL(x_{i':})} \sum_{j=1}^{p} \bar{b}_{x_i,x_{i'}} \bar{f}(x_{i:},x_{i':}) \hphantom{x}\mathbbm{1} \big[(x_{i:})\cancel{CL}(x_{i':})\big] \bigg) \bigg\}
\end{align}	

Adding the metric learning to the PCKM objective function and assuming that in equation (\ref{pckmeans}) the cost of violating constraints is equal to 1  results in the objective function of the MPCKM algorithm (\ref{mpckmeans}) \cite{bilenko2004integrating}:

\begin{align}\label{mpckmeans}
\CJ_{mpckm} &= \sum_{k=1}^{K} \sum_{\binom{i=1}{x_{i:} \in c_k}}^{n_k} \bigg( \sum_{j=1}^{p} a_{j}(x_{ij}-m_{kj})^2 - \sum_{j=1}^{p}log(a_{j}) + 
\nonumber\\&
\sum_{(x_{i:})ML(x_{i':})} \sum_{j=1}^{p} a_{j}f(x_{i:},x_{i':}) \hphantom{x}\mathbbm{1} \big[(x_{i:})\cancel{ML}(x_{i':})\big] + 
\nonumber\\&
\sum_{(x_{i:})CL(x_{i':})} \sum_{j=1}^{p} a_{j}\bar{f}(x_{i:},x_{i':}) \hphantom{x} 
\mathbbm{1} \big[(x_{i:})\cancel{CL}(x_{i':})\big] \bigg)
\end{align}	
where $a_{j}$ is a weight that parameterizes the Euclidean distance on the j-th dimension and $\sum_{j=1}^{p}log(a_{j})$ is a normalization constant \cite{xing2003distance} not allowing the weights to grow too large. 

An iterative algorithm for minimizing the function (\ref{mpckmeans}) is given by the algorithm below, see also \cite{bilenko2004integrating}:

\begin{enumerate}
	\item Initialise $K$ initial centroids $M = \{m_{1j}, \dots, m_{Kj}\}$ using an initialisation method and $W$ as a diagonal matrix with values $w_{1} = \dots = w_{p} = 1$. 
	\item Assign each data point to cluster $k^*$ so that,
	\begin{align}\label{mpckmeansobj}
	k^* &= \underset{k}{argmin}\bigg\{ \sum_{\binom{i=1}{x_{i:} \in c_k}}^{n_k} \bigg( \sum_{j=1}^{p} a_{j}(x_{ij}-m_{kj})^2 - \sum_{j=1}^{p}log(a_{j}) + 
	\nonumber\\&
	\sum_{(x_{i:})ML(x_{i':})} \sum_{j=1}^{p} a_{j}f(x_{i:},x_{i':}) \hphantom{x}\mathbbm{1} \big[(x_{i:})\cancel{ML}(x_{i':})\big] + 
	\nonumber\\&
	\sum_{(x_{i:})CL(x_{i':})} \sum_{j=1}^{p} a_{j}f(x_{i:},x_{i':}) \hphantom{x}
	\mathbbm{1} \big[(x_{i:})\cancel{CL}(x_{i':})\big] \bigg) \bigg\}
	\end{align}	
	
	\item Recompute the cluster centroids by taking the mean of the data points assigned to them, i.e for the k-th cluster, if it contains $n_k$ elements its centroid is computed as $m_{kj} = \frac{1}{n_k}\sum_{i=1}^{n_k}x_{ij}, \forall x_{ij}\in c_k \forall j$.
	
	\item Update the weights $\forall j$,
	\begin{align}\label{mpckmeansobj2}
	&a_{j} = n_k \Bigg( \sum_{k=1}^{K} \sum_{\binom{i=1}{x_{i:} \in c_k}}^{n_k} \bigg( (x_{ij}-m_{kj})^2 + 
	\nonumber\\&
	\sum_{(x_{i:})ML(x_{i':})} a_{j} f(x_{i:},x_{i':}) \hphantom{x}\mathbbm{1} \big[(x_{i:})\cancel{ML}(x_{i':})\big] + 
	\nonumber\\&
	\sum_{(x_{i:})CL(x_{i':})} a_{j}\bar{f}(x_{i:},x_{i':}) \hphantom{x} 
	\mathbbm{1} \big[(x_{i:})\cancel{CL}(x_{i':})\big] \bigg) \Bigg )^{-1}
	\end{align}			
	
	\item Iterate through steps 2 to 4 until the convergence. Various criteria can be used for convergence e.g. maximum number of iteration reached or minimum changes in the objective function.
	
\end{enumerate}	
The algorithm returns the final clusters (centroids and element assignments). The weights correspond to the  learnt metric that shapes the feature space accordingly to satisfy the input constraints. 

\subsection{Adding constraints to the Sparse K-means algorithm}

We extend the sparse clustering algorithm (SKM) proposed by \cite{witten2010framework} so that it can also take advantage of pairwise constrains. This extended version, which we name Pairwise Constrained Sparse K-Means (PCSKM), aims to maximise the objective function in equation (\ref{pcskmeans})

\begin{align}\label{pcskmeans}
J_{pcskmeans} =& \sum_{j=1}^{p} w_{j} \Bigg[ \sum_{i=1}^{n}(x_{ij}-{\mu}_{1j})^2 - 
\Bigg(\sum_{k=1}^{K}\sum_{\binom{i=1}{x_{i:} \in c_k}}^{n_k}(x_{ij}-m_{kj})^2 + 
\nonumber\\&
\sum_{(x_{i:})ML(x_{i':})}f(x_{i:},x_{i':})\hphantom{x}\mathbbm{1} \big[(x_{i:})\cancel{ML}(x_{i':})\big] +
\nonumber\\&
\sum_{(x_{i:})CL(x_{i':})} \bar{f}(x_{i:},x_{i':}) \hphantom{x}\mathbbm{1} \big[(x_{i:})\cancel{CL}(x_{i':})\big] \Bigg)
\Bigg]  \\ \nonumber
\text{subject to} & \hphantom{xxxxx}
\sum_{j=1}^{p} w_{j}^{2} \leq 1 \text{,}\hphantom{x} \sum_{j=1}^{p} \abs{w_{j}} \leq s \text{,}\hphantom{x} w_{j} \geq 0\hphantom{x} \forall j 
\end{align}	

Based on \cite{witten2010framework} this problem can be written in the form of 
\begin{align}\label{pcskmeansoptt}
&\underset{w_j}{maximize} \bigg\{ \sum_{j=1}^{p} w_{j}\gamma'_j \bigg\} \hphantom{x,} \forall j\nonumber \\
&\text{subject to} \hphantom{xxxxx} \sum_{j=1}^{p} w_{j}^{2} \leq 1 \text{,}\hphantom{x} \sum_{j=1}^{p} \abs{w_{j}} \leq s \text{,}\hphantom{x} w_{j} \geq 0\hphantom{x}
\end{align}	
where,
\begin{align}\label{pcskmeansoptt2}
\gamma'_j = &\sum_{i=1}^{n} (x_{ij}-{\mu}_{1j})^2 - \sum_{k=1}^{K}\sum_{\binom{i=1}{x_{i:} \in c_k}}^{n_k}(x_{ij}-m_{kj})^2
\nonumber\\&
\sum_{(x_{i:})ML(x_{i':})} f(x_{i:},x_{i':}) \hphantom{x}\mathbbm{1} \big[(x_{i:})\cancel{ML}(x_{i':})\big] +
\nonumber\\&
\sum_{(x_{i:})CL(x_{i':})}  \bar{f}(x_{i:},x_{i':}) \hphantom{x}\mathbbm{1} \big[(x_{i:})\cancel{CL}(x_{i':})\big]
\end{align}		
Similar to Sparse K-Means we define the following algorithm:

\begin{enumerate}
	\item Initialise $K$ initial centroids $M = \{m_{1j}, \dots, m_{Kj}\}$ using an initialisation method and set $w_{1} = \dots = w_{p} = \frac{1}{\sqrt{p}}$.
	\item Holding the weights fixed, maximize (\ref{pcskmeans}) with respect to $M$. This can be implemented by performing a variant of the K-Means algorithm where the point assignment to the nearest cluster is given by:
	\begin{align}\label{lpckmeansobj2}
	&k_{pcskm}^* = \underset{k}{argmin}\bigg\{ \sum_{\binom{i=1}{x_{i:} \in c_k}}^{n_k} \bigg( \sum_{j=1}^{p} w_j(x_{ij}-m_{kj})^2 + 
	\nonumber\\&
	\sum_{(x_{i:})ML(x_{i':})} \sum_{j=1}^{p}w_j f(x_{i:},x_{i':}) \hphantom{x}\mathbbm{1} \big[(x_{i:})\cancel{ML}(x_{i':})\big] + 
	\nonumber\\&
	\sum_{(x_{i:})CL(x_{i':})} \sum_{j=1}^{p}w_j \bar{f}(x_{i:},x_{i':}) \hphantom{x}\mathbbm{1} \big[(x_{i:})\cancel{CL}(x_{i':})\big] \bigg) \bigg\}
	\end{align}	
	
	\item Step 3 of Sparse K-Means replacing $\gamma_j$ with $\gamma_{j'}$. 
	
	\item Step 4 of the Sparse K-Means algorithm.
\end{enumerate}				

We note that K-Means and Sparse K-Means are element order invariant meaning that with the same initialisation method (and the same random seed if the centroids initialisation method is stochastic) it will produce the same results. This is true for PCSKM but only if the elements are processed in the same order due to the constraints. We find that this does not effect the clustering performance that we report.

\section{Results}
Our benchmark includes the real world data sets fisheriris, ionosphere and digits from the UCI repository \cite{asuncion2007uci} which have unknown feature quality. From the original digits data set we created 10 different sub-sets by randomly sampled elements of three classes, 0-4-8 and 3-8-9, 50 elements per class (3-8-9 digits were also used in the study of \cite{bilenko2004integrating}). We also added two synthetic data sets that were generated based on \cite{brodinova2017robust} which we used in our previous study \cite{vouros2019empirical}. These synthetic data sets consist of known informative and uninformative features without noise. More specifically, they consist of 120 10-dimensional data points spread equally across 3 clusters. The first set has 5 informative and 5 uninformative features while the second 3 informative and 7 uninformative features. Finally, we used two reduced data sets from the  studies of \cite{gehring2015detailed,vouros2018generalised} that consist of $8$ features of unknown importance that describe rodent path segments in the Morris Water Maze experiment. The first set (named TT-SC-ST) contains a total of $424$ data points and $3$ classes (Thigmotaxis: $168$, Scanning: $182$ and Scanning Target: $74$ data points); the second set (named TT-CR-ST) contains a total of $406$ data points and $3$ classes (Thigmotaxis: $168$, Chaining Response: $164$ and Scanning Target: $74$ data points). The dimensionality of both sets was increased with the addition of $10$ features. The $9^{th}$ feature is the path segment length which is uninformative given the fact that all the segments were created to have approximately the same length and the next $9$ features were generated from random shuffling of the segment length feature.

In our first experiment we assessed the performance of PCSKM versus other unsupervised algorithms (LKM and SKM) and semi-supervised algorithms (PCKM and MPCKM). We tested the performance of these algorithms using the deterministic initialisation techniques of DKM++, ROBIN and Maximin. We also tested the semi-supervised algorithms with different number and types of constraints: only MUST-LINK, only CANNOT-LINK, and a random selection from both MUST-LINK and CANNOT-LINK. Finally we tested the algorithms with the seeding initialisation method considering only a random selection from both MUST-LINK and CANNOT-LINK constraints since by construction it requires MUST-LINK constraints but performs best with both (see \cite{bilenko2004integrating}).

\begin{figure*}
	\centering
	\includegraphics[width=0.85\linewidth]{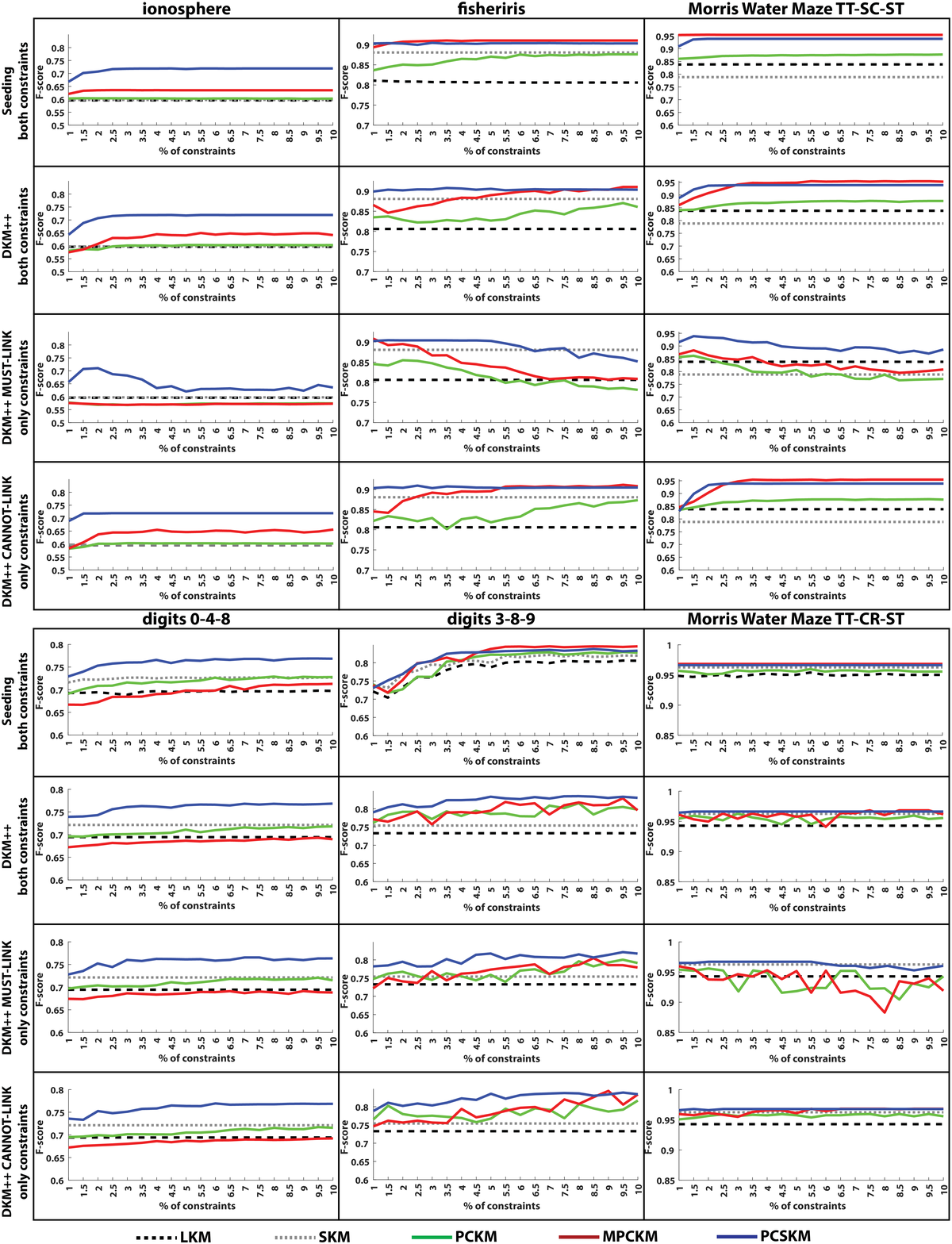}
	\caption{\textbf{Performance of PCSKM versus other unsupervised and semi-supervised algorithms.} 
		Each row compares the algorithms on six data sets (ionosphere, fisheriris, digits 0-4-8 and 3-8-9 and Morris Water Maze TT-SC-ST and TT-CH-ST) using different types of constrains. First row (Seeding both constraints): we used Seeding for cluster initialisation and a random selection from all the constraints, both MUST-LINK and CANNOT-LINK. Second row (DKM++ both constraints): similar as before but DKM++ initialisation was used. Third and fourth rows (MUST-LINK, CANNOT-LINK): we used DKM++ initialisation and a random selection of only MUST-LINK or CANNOT-LINK. For the SKM and PCSKM the sparsity value with the best performance was selected. For the ionosphere and digits 0-4-8 data, our algorithm has a clear advantage compared with the other methods. For the fisheriris, digits 3-8-9 and Morris Water Maze TT-SC-ST data, cluster initialisation with Seeding offers an advantage to the MPCKM algorithm compared to the DKM++ initialisation method. For all the data sets apart from the digits, using only MUST-LINK constraints has a negative effect. Clearly, the type of constraints can greatly affect the clustering performance while the initialisation method has less effect apart from the case of the MPCKM algorithm. PCSKM is in general more robust to initial conditions and its performance either surpass or is close to the performance of the other algorithms. 
	} \label{variousModels}  
\end{figure*}

For this experiment we show the results on the fisheriris, ionosphere, digits and the Morris Water Maze data sets since all the algorithms, as expected, performed equally well on the synthetic data due to their distinctive clusters. Figure \ref{variousModels} shows the results of the performance comparison. Our algorithm (PCSKM) was designed for its feature selection property rather than its performance and yet we demonstrate that it has better performance in most cases and, where its performance is worse, the difference to the best performing algorithm is relatively small. Interestingly the use of only MUST-LINK constraints has a negative effect in the performance of all the semi-supervised algorithms except for the digits data sets. Using either ROBIN or Maximin initialisation leads to similar conclusions (results in appendix). 

The performance was tested using a similar evaluation methodology as in \cite{basu2004active,bilenko2004integrating}. We run 10-fold cross validation using all the data but splitting the labels into training and test sets. The performance on each fold was assessed based on the F-score, an information retrieval measure, adapted for evaluating clustering by considering same-cluster pairs, see also \cite{bilenko2004integrating}. The clustering algorithm was run on the whole data set, but the F-score was calculated only on the test set. Constraints were created from only the training set. To evaluate the number of constraints required for achieving good performance, we run the algorithms using $1\%$ to $10\%$ of the total constraints generated from the training labels of each fold (Table \ref{datastats}). Results were averaged over 25 runs of 10 folds, each run with a random selection of constraints. Specifically for the two digits data sets 0-4-8 and 3-8-9 we calculated the average performance over the 10 sub-sets that we have created. 

In all experiments, the number of clusters $K$ is set to be equal to the number of classes as in other studies, e.g. \cite{bilenko2004integrating}. Sparse algorithms (i.e. SKM and PCSKM) have one more parameter that needs tuning, the sparsity parameter $s$. We selected a value for $s$ in the following way: we run the clustering process for $s=1.1$ to $s=\sqrt{p}$ with a step of $0.2$, where $p$ is the dimensionality of the data set, similar to \cite{brodinova2017robust}. Among these $s$ values we selected the one that yields the best value of F-score. 

We quantified the overall performance of PCSKM versus other algorithms. For each initialisation method (Seeding, DKM++, ROBIN, Maximin) constraints type (both, MUST-LINK only and CANNOT-LINK only) and data set (6 data sets in total) we calculated the average performance over the number of constraints (60 cases in total, since Seeding has only both constraints) per algorithm (PCSKM, PCKM, MPCKM, KM, SKM). Afterwards we used the Paired Samples Wilcoxon Test hypothesizing that the performances between any algorithm and PCSKM have the same distributions with the same medians. A p-value less than 0.05 in our analyses, lead us to discard the null hypothesis. Based on the results, there is significant difference between PCSKM and KM ($p<1^{-10}$), SKM ($p<1^{-10}$), PCKM ($p<1^{-10}$), MPCKM ($p<0.001$) in favor of the PCSKM.

We also quantified the performance difference among the different types of constraints (both, MUST-LINK only and CANNOT-LINK only) in a similar manner as we did with the overall performance. For each initialisation method (DKM++, ROBIN, Maximin), semi-supervised algorithm (PCKM, MPCKM, SKM, PCSKM) and data sets (6 in total excluding the synthetic), we calculated the average performance over the number of constraints (72 cases in total). Afterwards we used the Paired Samples Wilcoxon Test setting again a p-value less than 0.05 to discard the null hypothesis. Based on the results, there is significant difference between MUST-LINK only and CANNOT-LINK only constraints in favor of the latter (p-value $< 0.01$) and there is significant difference between MUST-LINK only and both constraints in favor of the latter (p-value $< 0.001$). No significant difference was detected between the CANNOT-LINK only and both constraints (p-value $> 0.1$).

	\begin{table*}[!t]
	\centering
	\caption{\textbf{Data set constraints.} Column {\it Points} shows the total amount of data points of each data set; {\it CV10 labels} shows the number of labels that can be used for training (90\% of the Points since we use 10-fold cross validation); {\it CV10 constraints} shows the total number of constraints generated from these labels, i.e. the size of the pool. The last column refers to the number of constraints (minimum and maximum) that were randomly sampled from the pool and used in the training. The minimum corresponds to the 1\% and the maximum to the 10\% of CV10 constraints.}
	\label{datastats}
	\begin{tabular}{|l|c|c|c|c|}
		\hline
		\multicolumn{1}{|c|}{\textbf{Data set}} & \textbf{Points} & \textbf{CV10 labels} & \textbf{CV10 constraints} & \textbf{\begin{tabular}[c]{@{}c@{}}CV10 constraints\\ {[}1\% , 10\%{]}\end{tabular}} \\ \hline
		\textbf{fisheriris}                     & 150             & 135                & 9045                      & {[}90 , 905{]}                                                                       \\ \hline
		\textbf{ionosphere}                     & 351             & 316                & 49738                     & {[}497 , 4974{]}                                                                     \\ \hline
		\textbf{digits 0-4-8 and 3-8-9}                            & 150                & 135                     & 9045                          & {[}90 ,  905{]}                                                                                     \\ \hline
		\textbf{Morris Water Maze TT-SC-ST}                            & 424                & 382                     & 72618                          & {[}726 ,  7262{]}                                                                                     \\ \hline
		\textbf{Morris Water Maze TT-CH-ST}                            & 406                & 365                     & 66576                          & {[}666 ,  6658{]}                                                                                     \\ \hline
		\textbf{Brodinova (2 sets)}                      & 120             & 108                 & 5778                      & {[}58 , 578{]}                                                                       \\ \hline
	\end{tabular}
\end{table*}

\begin{figure*}
	\centering
	\includegraphics[width=\linewidth]{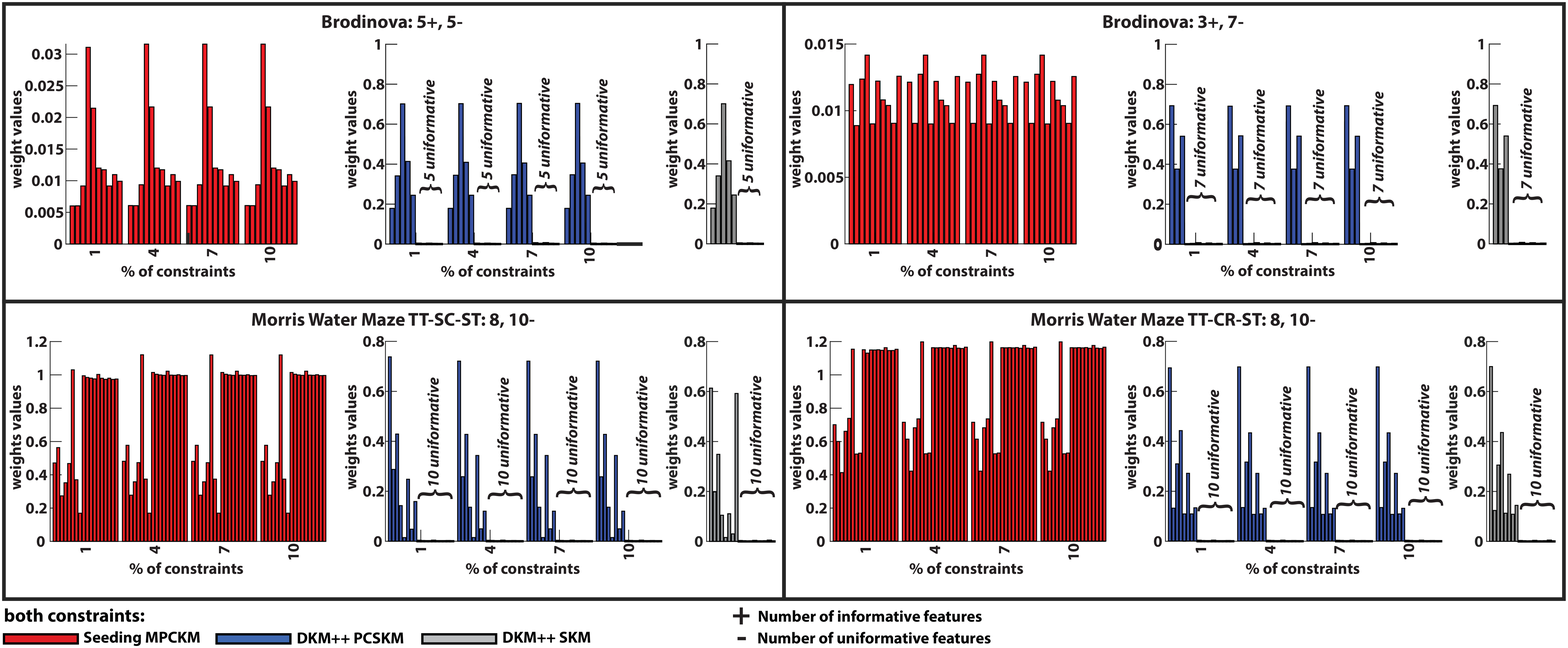}
	\caption{\textbf{PCSKM feature selection capabilities versus other algorithms in synthetic and the Morris Water Maze data sets.} Each bar plot shows the weight values of the features versus the number of constraints. Red bars: MPKM (semi-supervised), blue bars PCSKM (semi-supervised), gray bars: SKM (unsupervised). For SKM and PCSKM the bars show the average weight value of a feature over different sparsity ($\boldsymbol{s}$) values (from $\boldsymbol{s = 1.1}$ to $\boldsymbol{s = \sqrt{p}}$, where $\boldsymbol{p}$ is the dimensionality of the data set, with step 0.2). The $\boldsymbol{+}$ and $\boldsymbol{-}$ signs indicate the number of informative and uninformative features (uninformative features are always plotted last). In the case of the Morris Water Maze the quality of the first 8 features is unknown but the last 10 are uninformative. The SKM and PCSKM correctly identifies the known uninformative features regardless of the choice of the $\boldsymbol{s}$ parameter in all the cases. The feature selection mechanism of the PCSKM is not affected by the number of constraints. The weights of the MPCKM algorithm are not indicative of the feature quality and in all the cases the algorithm uses the uninformative features. In the plots we show only the case when both type of constraints are used but we observe the same result for the other constraint types cases regardless of the initialisation method.} \label{MWMweights}  
\end{figure*}

In our second experiment, we assess the feature selection capabilities of our algorithm. We used the two synthetic data sets for which we have knowledge of both the informative and the uninformative features as well as the Morris Water Maze data sets for which we have knowledge only of the uninformative features. In Figure \ref{MWMweights} we compare the feature selection capabilities of MPCKM, SKM and our PCSKM, and the results are consistent for all initialisation methods we tested, including ROBIN and Maximin (results in appendix). For SKM and PCSK we show average results over different values of $s$ from $1.1$ to $\sqrt{p}$, where $p$ is the dimensionality of the data set, with step $0.2$ as in \cite{brodinova2017robust}. This is to demonstrate that the feature selection does not strongly depend on optimally selecting the value of $s$. The weight values are plotted as a function of number of constraints (where applicable) demonstrating that the feature selection capabilities of PCSKM are not affected by it. We also contaminated the digits data sets with 4 uninformative features generated from exponential distributions (results in appendix) and again the PCSKM algorithm was able to assign a 0 weight to them. On the contrary, the MPCKM algorithm, which learns a metric, makes use of uninformative features, see Figure \ref{MWMweights} Seeding MPCKM. In addition, in both contaminated digit sets the uninformative features have on average higher weights that the original features, and this is true across all initialisation methods (Seeding, DKM++, ROBIN, Maximin).  

\section{Discussion}
We modified an existing clustering algorithm with a feature selection mechanism to address semi-supervised problems where few labels are available. We hypothesized that the ability of this algorithm in detecting informative and non-informative features is better than an alternative semi-supervised algorithm with metric learning, while its performance will be at least equal. We also demonstrated that the constraints improve the performance of our algorithm on classification problems in comparison to its unsupervised version.

We tested our modified algorithm (PCSKM) under different initial conditions. We have found that its performance is almost equivalent to other semi-supervised algorithms and in many cases better, regardless of the initialisation method (DKM++, MPCKM, ROBIN or Maximin), type (MUST-LINK only, CANNOT-LINK only, both) or number of constraints. We also showed that its feature selection mechanism is robust to the initialisation method and the number of constraints, and its weight assignments can be used to indicate informative or uninformative features in all the data sets we tested. On the contrary the MPCKM algorithm, which learns a metric, while it performs best or in par with PCSKM when the Seeding initialisation method is used (with the exception of the data sets ionosphere and digits 0-4-8), it makes use of the uninformative features. As a consequence the learned weights cannot be used to evaluate the quality of the features.

We have also showed that the performance of semi-supervised algorithms can be affected by initialisation procedures, similar to unsupervised methods \cite{vouros2019empirical}, and by the type of constraints. By experimenting with different number of constraints, different initialisation methods and different semi-supervised algorithms we draw conclusions about the goodness of each constraints type. Only MUST-LINK constraints have a negative effect on the performance for all the data sets we tested (apart from the digits). We speculate that this may be the case because data points belonging to the same class do not necessarily belong to the same cluster, while CANNOT-LINK constraints are more informative: data points belonging to different classes should not belong to the same cluster. Nevertheless, PCSKM could cope with the MUST-LINK constraints much better than MPCKM and PCKM.

In addition, the Seeding initialisation method proposed for the MPCKM in the study of \cite{bilenko2004integrating} has mostly a positive effect on that particular algorithm. In the case of Morris Water Maze data sets, the initial 8 features were also engineered to perform best with the MPCKM algorithm \cite{gehring2015detailed}. Nevertheless, our algorithm maintains a close performance to MPCKM and based on our hypothesis testing, it has on average the best overall performance. This is, perhaps, due to the use of quality features that should have in general a positive effect on the overall clustering performance.

Finally, our proposed PCSKM algorithm, similar to SKM, requires an additional sparsity (\textit{s}) parameter. In the study of \cite{brodinova2017robust} both (\textit{s}) and the number of clusters $K$ are chosen using the gap statistic. In our case, due to the semi-supervised setting, we can set $K$ equal to the number of classes and use the F-score to select the sparsity parameter $s$ avoiding the computationally more expensive gap criterion.

\bibliographystyle{unsrt}  
\bibliography{template}  
\cleardoublepage

\appendix

\section{Appendix}

\begin{figure*}[h]
	\centering
	\includegraphics[width=0.85\linewidth]{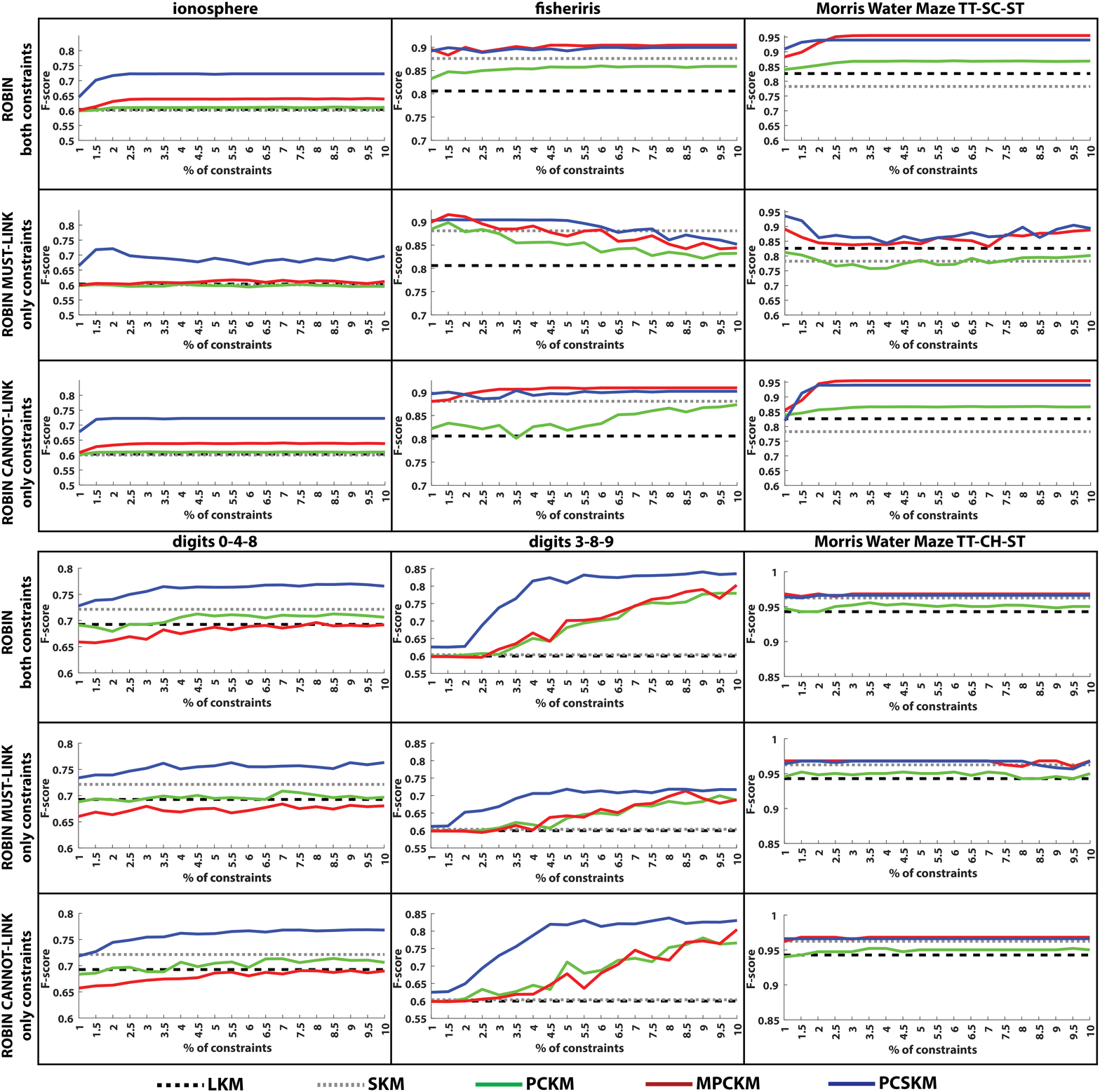}
	\caption{\textbf{Performance of PCSKM as opposed to other unsupervised and semi-supervised algorithms using the ROBIN initialisation method.} Each row compares the algorithms over six data sets (ionosphere, fisheriris, digits 0-4-8 and 3-8-9 and MorrisWater Maze TT-SC-ST and TT-CH-ST) using different
	types of constrains. First row (ROBIN both constraints): ROBIN was used for clustering initialisation and there has been a random selection from all the constraints, both MUST-LINK and CANNOT-LINK. Second and third rows (MUST-LINK, CANNOT-LINK): ROBIN initialisation was used and there was a random selection of only MUST-LINK or CANNOT-LINK. For the SKM and PCSKM the sparsity value with the best performance was selected.
	} \label{sup1}  
\end{figure*}

\begin{figure*}[h]
	\centering
	\includegraphics[width=0.85\linewidth]{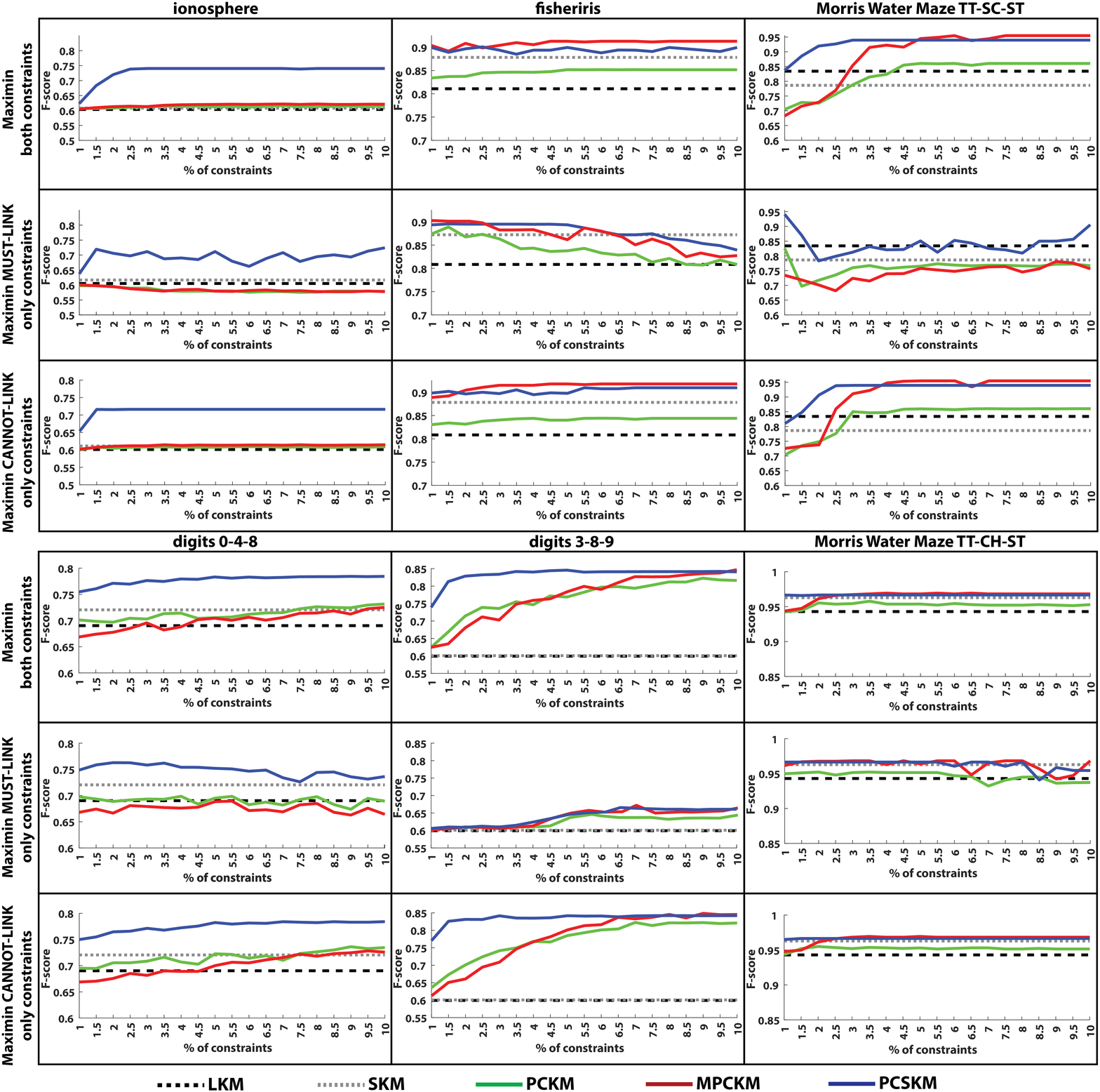}
	\caption{\textbf{Performance of PCSKM as opposed to other unsupervised and semi-supervised algorithms using the Maximin initialisation method.} Each row compares the algorithms over six data sets (ionosphere, fisheriris, digits 0-4-8 and 3-8-9 and Morris Water Maze TT-SC-ST and TT-CH-ST) using different types of constrains. First row (ROBIN both constraints): ROBIN was used for clustering initialisation and there has been a random selection from all the constraints, both MUST-LINK and CANNOT-LINK. Second and third rows (MUST-LINK, CANNOT-LINK): Maximin initialisation was used and there was a random selection of only MUST-LINK or CANNOT-LINK. For the SKM and PCSKM the sparsity value with the best performance was selected.
	} \label{sup2}  
\end{figure*}

\begin{figure*}[h]
	\centering
	\includegraphics[width=\linewidth]{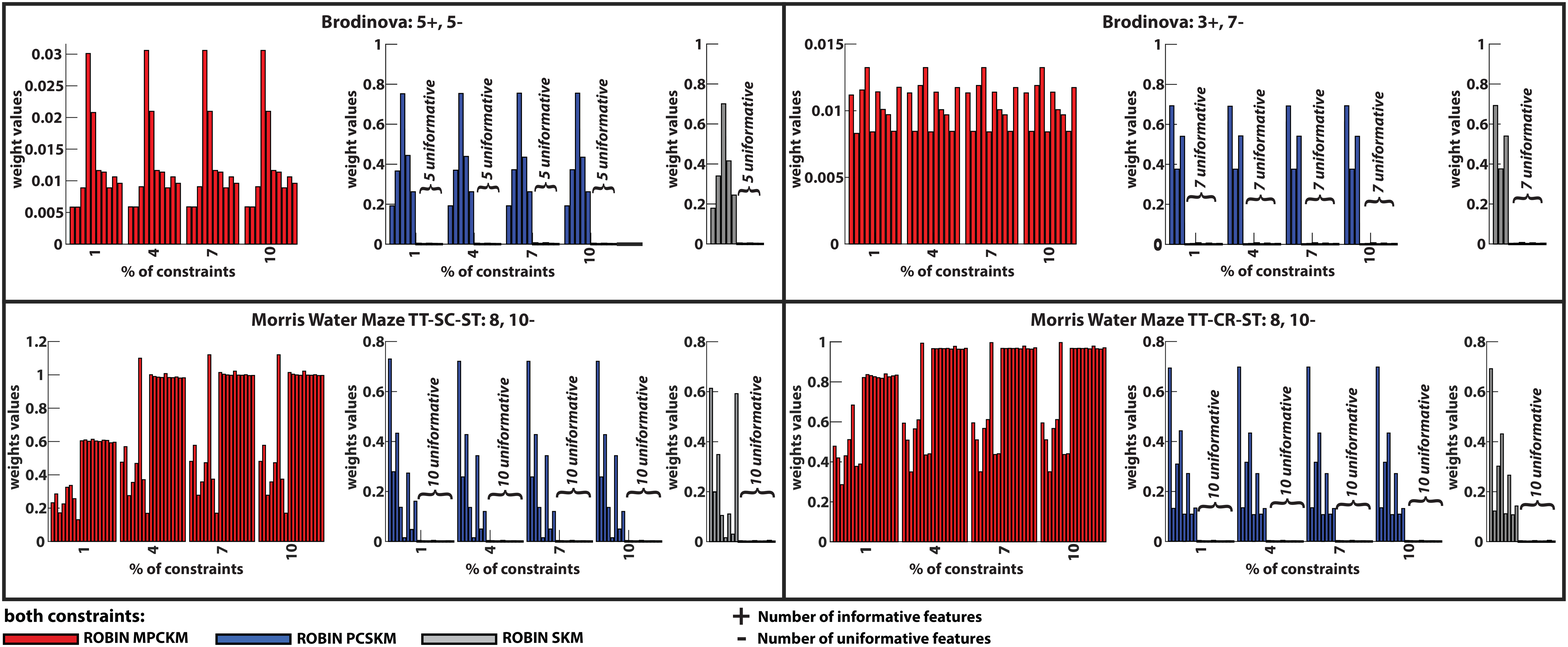}
	\caption{\textbf{PCSKM feature selection capabilities as opposed to other algorithms in synthetic and the Morris Water Maze data sets using the ROBIN initialisation method.} Each bar plot shows the value of each one of the features of the data set over the number of constraints. Red bars: MPKM (semisupervised), blue bars PCSKM (semi-supervised), gray bars: SKM (unsupervised). Red bars: MPKM (semi-supervised), blue bars PCSKM (semi-supervised), gray bars: SKM (unsupervised). For SKM and PCSKM the bars show the average weight value of a feature over different sparsity ($\boldsymbol{s}$) values (from $\boldsymbol{s = 1.1}$ to $\boldsymbol{s = \sqrt{p}}$, where $\boldsymbol{p}$ is the dimensionality of the data set, with step 0.2). The $\boldsymbol{+}$ and $\boldsymbol{-}$ signs indicate
	the number of informative and uninformative features (uninformative features are always plotted last). In the case of the Morris Water Maze the quality of the first 8 features is unknown but the last 10 are uninformative. The SKM and PCSKM correctly identifies the known uninformative features regardless of the $s$ parameter value in all the cases. Specifically for the PCSKM the feature selection mechanism is not affected by the constraints. The MPCKM algorithm fails to show any indication about the feature quality based on the feature weights and in all the cases it uses the uninformative features. In the plots we show only the case when both type of constraints are used but we observe the same result for the other constraint types cases regardless of the used initialisation method.} \label{sup4}  
\end{figure*}

\begin{figure*}[h]
	\centering
	\includegraphics[width=\linewidth]{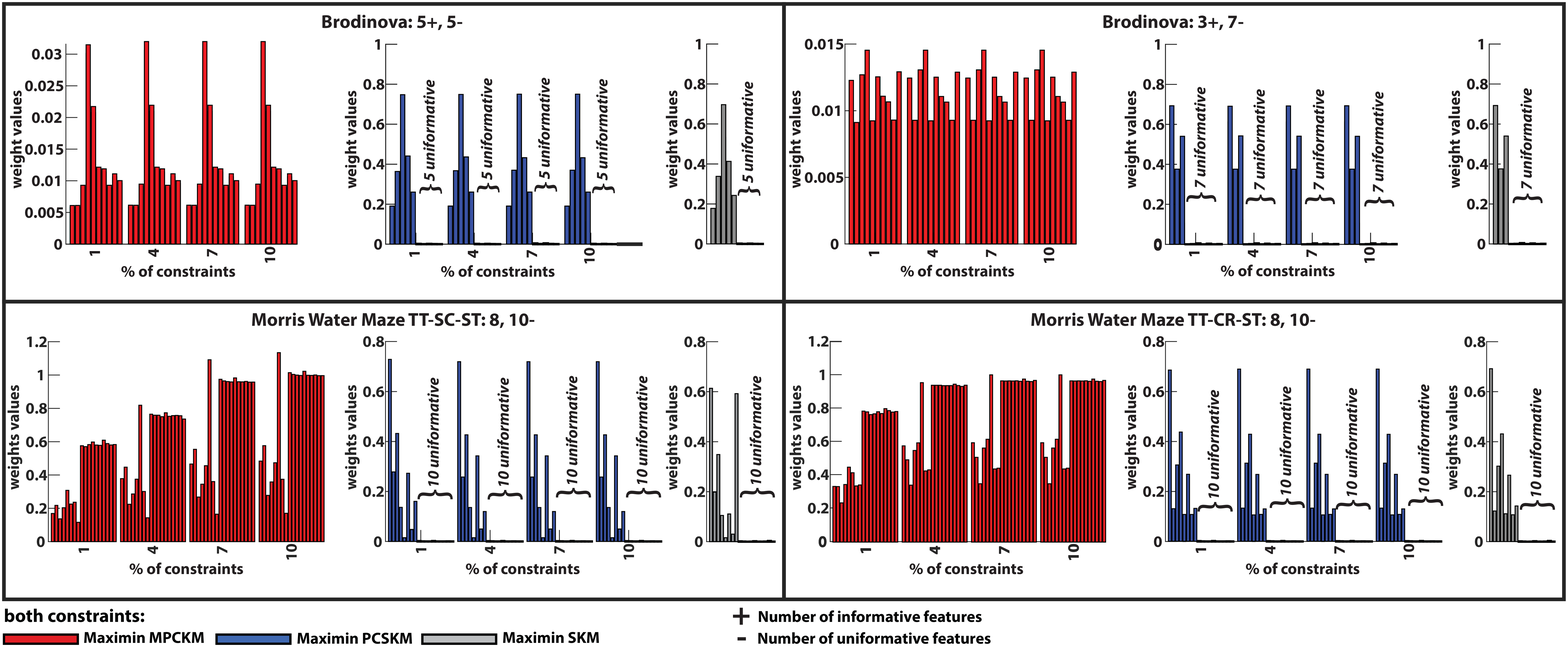}
	\caption{\textbf{PCSKM feature selection capabilities as opposed to other algorithms in synthetic and the Morris Water Maze data sets using the Maximin initialisation method.} Each bar plot shows the value of each one of the features of the data set over the number of constraints. Red bars: MPKM (semisupervised), blue bars PCSKM (semi-supervised), gray bars: SKM (unsupervised). Red bars: MPKM (semi-supervised), blue bars PCSKM (semi-supervised), gray bars: SKM (unsupervised). For SKM and PCSKM the bars show the average weight value of a feature over different sparsity ($\boldsymbol{s}$) values (from $\boldsymbol{s = 1.1}$ to $\boldsymbol{s = \sqrt{p}}$, where $\boldsymbol{p}$ is the dimensionality of the data set, with step 0.2). The $\boldsymbol{+}$ and $\boldsymbol{-}$ signs indicate the number of informative and uninformative features (uninformative features are always plotted last). In the case of the Morris Water Maze the quality of the first 8 features is unknown but the last 10 are uninformative. The SKM and PCSKM correctly identifies the known uninformative features regardless of the $s$ parameter value in all the cases. Specifically for the PCSKM the feature selection mechanism is not affected by the constraints. The MPCKM algorithm fails to show any indication about the feature quality based on the feature weights and in all the cases it uses the uninformative features. In the plots we show only the case when both type of constraints are used but we observe the same result for the other constraint types cases regardless of the used initialisation method.} \label{sup5}  
\end{figure*}

\begin{figure*}[h]
	\centering
	\includegraphics[width=\linewidth]{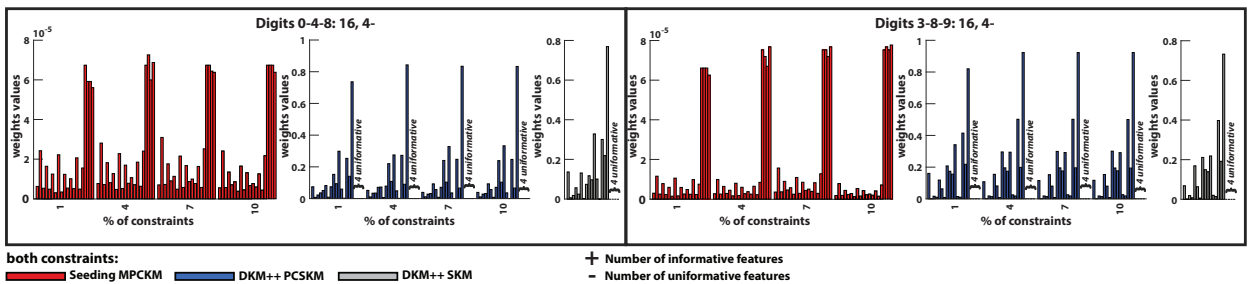}
	\caption{\textbf{PCSKM feature selection capabilities as opposed to other algorithms in the digits data sets contaminated with 4 uninformative features.}} \label{sup6}  
\end{figure*}

\end{document}